\pdfoutput=1
\documentclass[11pt]{article}
\usepackage{acl}
\usepackage{times}
\usepackage{latexsym}
\usepackage[T1]{fontenc}
\usepackage[utf8]{inputenc}
\usepackage{microtype}
\usepackage{graphicx}
\usepackage{subfig}
\usepackage[inline]{enumitem}
\newcommand{\rt}[1]{\rotatebox{90}{#1}}
\usepackage{booktabs}
\usepackage{multirow}
\usepackage{tabularx}
\usepackage[hang,flushmargin]{footmisc}
\usepackage{xspace}
\usepackage{scalefnt}
\newcommand{\F}{F$_1$\xspace}

\newcommand{\vertimg}[2]{\raisebox{-.7\height}{\includegraphics[scale=#1]{#2}}}

\title{On the Complementarity of Images and Text for the\\ Expression of Emotions in Social Media}

\author{%
  Anna Khlyzova$^{1,2}$, Carina Silberer$^{2}$ \and Roman Klinger$^{2}$ \\
  $^{1}$Computer Science and Engineering, University of South Florida, USA\\
$^{2}$Institut f{\"u}r Maschinelle Sprachverarbeitung, University of Stuttgart, Germany\\
  \texttt{\{anna.khlyzova,carina.silberer,roman.klinger\}}\\\texttt{@ims.uni-stuttgart.de}
}
 
\begin{document}
\maketitle
\begin{abstract}
  Authors of posts in social media communicate their emotions and what
  causes them with text and images.  While there is work on emotion
  and stimulus detection for each modality separately, it is yet
  unknown if the modalities contain complementary emotion information
  in social media. We aim at filling this research gap and contribute
  a novel, annotated corpus of English multimodal Reddit posts.  On
  this resource, we develop models to automatically detect the
  relation between image and text, an emotion stimulus category and
  the emotion class. We evaluate if these tasks require both
  modalities and find for the image--text relations, that text alone
  is sufficient for most categories (complementary, illustrative,
  opposing): the information in the text allows to predict if an image
  is required for emotion understanding.  The emotions of anger and
  sadness are best predicted with a multimodal model, while text alone
  is sufficient for disgust, joy, and surprise.  Stimuli depicted by
  objects, animals, food, or a person are best predicted by image-only
  models, while multimodal models are most effective on art, events,
  memes, places, or screenshots.
\end{abstract}

\section{Introduction}
The main task in emotion analysis in natural language processing is
emotion classification into predefined sets of emotion categories, for
instance, corresponding to basic emotions \citep[fear, anger, joy,
sadness, surprise, disgust, anticipation, and trust,
][]{Ekman1992,Plutchik:1980emotions}. In psychology, emotions are
commonly considered a reaction to an event which consists of a
synchronized change of organismic subsystems, namely
neurophysiological changes, reactions, action tendencies, the
subjective feeling, and a cognitive appraisal
\citep{Scherer2001}. These theories recently received increasing
attention, for instance, by comparing the way how emotions are
expressed, based on these components \citep{Casel2021}, and by
modelling emotions in dimensional models of affect \citep{Buechel2017}
or appraisal \citep{Hofmann2020}. Further, the acknowledgment of
emotions as a reaction to some relevant event \citep{Scherer2005}
leads to the development of stimulus detection systems. This task is
formulated in a token-labeling setup
\citep[i.a.]{Song2015,Bostan2020,Kim2018,Ghazi2015,Oberlander2020}, as
clause classification
\citep[i.a.]{Gui2017,Gui2016,Gao2017,Xia2019a,Oberlander2020}, or as a
classification task into a predefined inventory of relevant stimuli
\citep{Mohammad2014}.

\begin{figure*}[t]
  \centering
  \urlstyle{same}
  \resizebox{0.9\linewidth}{!}{%
  \subfloat[][%
  joy/complementary/animal.\\ 
  \url{https://www.reddit.com/r/happy/comments/j76dog/my_everyday_joy_is_to_see_my_adorable_cat_smiles/}
  \label{fig:cat}
  ]{%
    \fbox{%
      \begin{minipage}[t][4.7cm]{0.28\linewidth}
      \sffamily\footnotesize\scalefont{0.8}
        My everyday joy is to see my adorable cat smiles. And I've just
        realized, my cat can "dance with music". Amazing!\\
        \includegraphics[height=3.8cm]{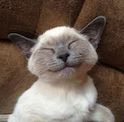}
      \end{minipage}
    }
  }
  \hspace{3mm}
  \subfloat[][%
  fear/complementary/animal.
  \url{https://www.reddit.com/r/WTF/comments/k2es5l/dont_move_to_australia_unless_you_can_handle/}%
  \label{fig:spider}
  ]{%
    \fbox{\hspace{2mm}%
      \begin{minipage}[t][4.7cm]{0.2\linewidth}
      \sffamily\footnotesize\scalefont{0.8}
        Don’t move to Australia unless you can handle these bad boys\\
        \includegraphics[height=3.8cm]{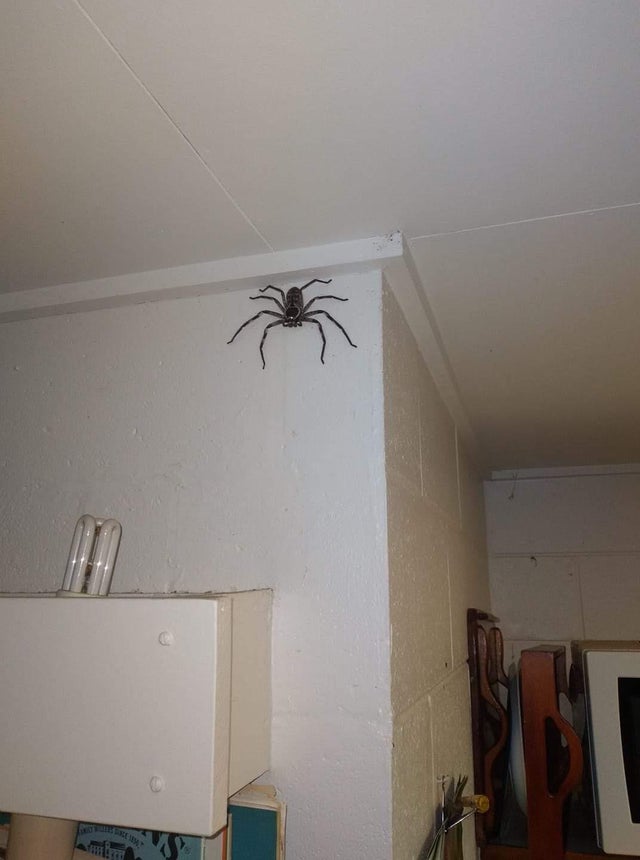}
      \end{minipage}\hspace{2mm}
    }
  }
  \hspace{3mm}
  \subfloat[][%
  surprise/complementary/object.\\
  \url{https://www.reddit.com/r/What/comments/exh0ms/why_didnt_it_fall/}
  \label{fig:bottle}
  ]{%
    \fbox{\hspace{2mm}%
      \begin{minipage}[t][4.7cm]{0.3\linewidth}
      \sffamily\footnotesize\scalefont{0.8}
        why didn’t it fall\\[2\baselineskip]
        \includegraphics[width=3.9cm]{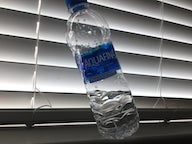}
      \end{minipage}
    }
  }
  }
  \caption{Example of posts from Reddit (annotation are
    emotion/relation/stimulus category).}
  \label{fig:example1}
\end{figure*}

In social media, users express emotions including text and
images. Most attention has been devoted to Twitter, due to its
easy-to-use API and popularity
\citep{Mohammad2012,Schuff2017,Wang2012b}. However, this platform has
a tendency to be text-focused, and has therefore not triggered too
much attention towards other modalities. Although text may be
informative enough to recognize an emotion in many cases, images may
modulate the meaning, or sometimes solely convey the emotion itself (see
examples in Figure~\ref{fig:example1}).  The growing popularity of
vision-centered platforms like TikTok or Instagram, and lack of
research on multimodal social media constitute a research gap.

With this paper, we study how users on social media make use of images
and text jointly to communicate their emotion and the stimulus of that
emotion. We assume that linking depictions of stimuli to the text
supports emotion recognition across modalities.  We study multimodal
posts on the social media platform
Reddit\footnote{\url{https://www.reddit.com/}}, given its wide
adoption, the frequently found use of images and text, and the
available programming interfaces to access the data
\citep{Baumgartner2020}.  Our goal is to understand how users choose
to use an image in addition to text, and the role of the relation, the
emotion, and the stimulus for this decision. Further we analyze if the
classification performance benefits from a joint model across
modalities. Figure~\ref{fig:example1} shows examples for Reddit
posts. In Figure~\ref{fig:cat}, both image and text would presumably
allow to infer the correct emotion even when considered in
isolation. In Figure~\ref{fig:spider}, additional knowledge of the
complementary role of the picture depicting an animal can inform an
emotion recognition model. In Figure~\ref{fig:bottle} the image alone
would not be sufficient to infer the emotion, but the text alone is.

We therefore contribute (1)~a new corpus of multimodal emotional posts
from Reddit, which is annotated for authors' emotions, image--text
relations, and emotion stimuli. We (2) analyze the relations of the
annotated classes and find that certain emotions are likely to appear
with certain relations and emotion stimuli. Further, we (3) use a
transformer-based language model \citep[pre-trained RoBERTa model,
][]{Liu:2019roberta} and a residual neural network \citep[Resnet50,
][]{He:2016resnet} to create classification models for the prediction
of each of the three classes mentioned above. We analyze for which
classification tasks multimodal models show an improvement over
unimodal models. Our corpus is publicly available at
\url{https://www.ims.uni-stuttgart.de/data/mmemo}.

\section{Related Work}

\noindent\textbf{Emotion Analysis.}
Emotion analysis has a rich history in various domains, such as fairy
tales \citep{alm-etal-2005-emotions}, email writing
\citep{Liu:2003email}, news headlines
\citep{strapparava-mihalcea-2007-semeval}, or blog posts
\citep{Mihalcea:2006blog, Aman:2008ER, Neviarouskaya2010}. The focus
of our study is on emotion analysis in social media, which has also
received considerable attention
\citep[i.a.]{purver-battersby-2012-experimenting,Wang2012b,Colneric:2018emotion,Mohammad2012,Schuff2017}.
Twitter\footnote{\url{https://www.twitter.com/}} is a popular social
media platform for emotion analysis, in both natural language
processing~(NLP) and computer vision. We point the reader to recent
shared tasks for an overview of the methods that lead to the current
state-of-the-art performance \citep{Klinger2018,Mohammad2018}.

One of the questions that needs to be answered when developing an
emotion classification system is that of the appropriate set of
emotions. There are two main theories regarding emotion models in
psychology that found application in NLP: discrete sets of emotions
and dimensional models. Psychological models that provide discrete
sets of emotions include Ekman's model of basic emotions \citep[anger,
disgust, surprise, joy, sadness, and fear,~][]{Ekman1992} and
Plutchik's wheel of emotions \citep[adding trust and
anticipation,~][]{Plutchik:1980emotions,
  Plutchik:2001emotions}. Dimensional models define where emotions lie
in a vector space in which the dimensions have another meaning,
including affect
\citep{Russell:1980circumplex,Bradley:1992remembering} and cognitive
event appraisal \citep{Scherer2005,Hofmann2020,Shaikh2009}. In our
study, we use the eight emotions from the Plutchik's wheel of
emotions.
    
\noindent\textbf{Multimodal Analyses.} 
The area of emotion analysis also received attention from the computer
vision community. A common approach is to use transfer learning from
general image classifiers \citep{He2019} or the analysis of facial
emotion expressions, with features of muscle movement
\citep{Silva1997} or deep learning
\citep{Li2020b}. \citet{Dellagiacoma2011} use texture and color
features to analyze social media content. Other useful properties of
images for emotion analysis include the occurrence of people, faces,
shapes of objects, and color distributions
\citep{ZhaoDingHuangetal2018,Lu2012}.

Such in-depth analyses are related to stimulus
detection. \citet{Peng2016} detect emotion-eliciting image
regions. They show, on a Flickr image dataset, that not only objects
\citep{Wu2020} and salient regions \citep{Zheng2017} have an impact on
elicited emotions, but also contextual background.  \citet{Yang2018},
inter alia, show that it is beneficial for emotion classification to
explicitly integrate visual information from emotion-eliciting
regions. Similarly, \citet{Fan2018} study the relationship between
emotion-eliciting image content and human visual attention.

\noindent\textbf{Image--Text Relation.}
A set of work aimed at understanding the
relation between images and text. \citet{Marsh:2003taxonomy} establish
a taxonomy of 49 functions of illustrations relative to text in US
government publications. The relations contain categories like
``elicit emotion'', ``motivate'', ``explains'', or ``compares'' and
``contrasts''. \citet{Martinec:2005taxonomy} aim at understanding both
the role of an image and of text. 

In contrast to these studies which did not develop machine learning
approaches, \citet{Zhang:2018taxonomy} develop automatic
classification methods for detection of relations between the image and a slogan in advertisments. They
detect if the image and the text make the same point, if one modality
is unclear without the other, if the modalities, when considered
separately, imply opposing ideas, and if one of the modalities is
sufficient to convey the message.  \citet{Weiland:2018taxonomy} focus
on detecting if captions of images contain complementary information.
\citet{vempala-preotiuc-pietro-2019-categorizing} infer relationship
categories between the text and image of Twitter posts to see how the
meaning of the entire tweet is
composed. \citet{kruk-etal-2019-integrating} focus on understanding
the intent of the author of an Instagram post and develop a hierarchy
of classes, namely \textsl{advocative}, \textsl{promotive},
\textsl{exhibitionist}, \textsl{expressive}, \textsl{informative},
\textsl{entertainment}, \textsl{provocative/discrimination}, and
\textsl{provocative/controversial}. They also analyze the relation
between the modalities with the classes \textsl{divergent},
\textsl{additive}, or \textsl{parallel}.  Our work is similar to the
two previously mentioned papers, as the detection which emotion is
expressed with a post is related to intent understanding.

\section{Corpus Creation}
To study the roles of images in social media posts, we create an
annotated Reddit dataset with labels of emotions, text--image
relations, and emotion stimuli. We first discuss our label sets and
then explain the data collection and annotation procedures.

\subsection{Taxonomies}
We define taxonomies for the emotion, relation, and stimulus tasks. 

\noindent\textbf{Emotion Classification.} To classify social media posts in
terms of what emotion the author likely felt when creating the post,
we use the Plutchik's wheel of emotions as the eight labels in our
annotation scheme, namely \textsl{anger}, \textsl{anticipation},
\textsl{joy}, \textsl{sadness}, \textsl{trust}, \textsl{surprise},
\textsl{fear}, and \textsl{disgust}.

\noindent\textbf{Relation Classification.}
To develop a classification scheme of relations of emotion-eliciting
image--text pairs, we randomly sampled 200 posts, and created a simple
annotation environment for preliminary annotation that displayed an
image--text pair next to questions to be answered (see Figure
\ref{fig:tool} in the Appendix). Based on the preliminary annotation,
we propose the following set of relation categories.
\begin{enumerate*}
\item \textsl{complementary}: the image is necessary to understand
the author's emotion; the text alone is not sufficient but when coupled
with the image, the emotion is clear;
\item \textsl{illustrative}: the image illustrates the text but the
text alone is enough to understand the emotion; the image does not
communicate the emotion on its own;
\item \textsl{opposite}: the image and the text pull in different
  directions; they are contradicting when taken separately, but when
  together, the emotion is clear;
\item \textsl{decorative}: the image is used for aesthetic purposes;
  the emotion is primarily communicated with the text while the image
  may seem unrelated;
\item \textsl{emotion is communicated with image only}: the text is
  redundant for emotion communication.
\end{enumerate*}

\begin{figure}[t]
  \centering
  \urlstyle{same}

  \subfloat[][%
  Relation: complementary.
  \url{https://www.reddit.com/r/sad/comments/jxgoxj/i_drew_this/} 
  ]{%
    \fbox{%
      \begin{minipage}[t][4cm]{0.4\linewidth}
       \sffamily\footnotesize\scalefont{0.8}
        I drew this\\
        \includegraphics[height=3.8cm]{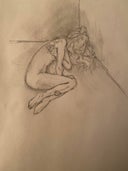}
      \end{minipage}
    }
  }
  \hfill
  \subfloat[][%
  Relation: illustrative.\\
  \url{https://www.reddit.com/r/happy/comments/jwje64/this_semester_has_kicked_me_in_a_way_none_other/}
  ]{%
    \fbox{%
      \begin{minipage}[t][4cm]{0.45\linewidth}
       \sffamily\footnotesize\scalefont{0.8}
        This semester has kicked me in a way none other has. Never
        cleaned my room until today. Forgot how big it could actually
        be. It’s the little things\\
        \includegraphics[height=2.5cm]{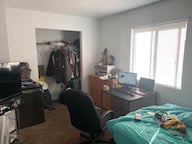}
      \end{minipage}
    }
  }
  \caption{Example of image--text relationships in posts.}
  \label{fig:example2}
\end{figure}

We show examples for the \textsl{complementary} and \textsl{illustrative} relations in
Figure~\ref{fig:example2}.
An example for the \textsl{opposite} relation could be an image with
an ugly creature with a text ``isn't he the prettiest thing in the world''.
Posts in which the text and the image are essentially unrelated fall
into the \textsl{decorative} category.
Posts where images have inspirational texts like ``No Happiness is
Ever Wasted'' and the text contains the same words would fall into the
last category (\textsl{image-only}).

\noindent\textbf{Stimulus Classification.}
Based on the preliminary annotation procedure described for the
relation taxonomy, we further obtain the following categories for emotion
stimuli in images of multimodal posts: \textsl{person/people},
\textsl{animal}, \textsl{object}, \textsl{food}, \textsl{meme},
\textsl{screenshot/text in image}, \textsl{art/drawing},
\textsl{advertisement}, \textsl{event/situation}, and \textsl{place}.
We provide examples of all stimuli in the Appendix in
Figure~\ref{fig:example3}.

\subsection{Data Collection}
We collect our multimodal data from Reddit, where posts are
published under specific subreddits, user-created areas of interest,
and are usually related to the topic of the group. Our data comes from
15 subreddits which we found by searching for emotion names. These
subreddits are ``happy'', ``happiness'', ``sad'', ``sadness'',
``anger'', ``angry'', ``fear'', ``disgusting'', ``surprise'',
``what'', ``WTF'', ``Cringetopia'', ``MadeMeSmile'', ``woahdude'',
which we complement by ``r/all''.

We collect the data from the Pushshift Reddit Dataset, a collection of
posts and comments from Reddit from 2015
\citep{baumgartner2020pushshift}, with the help of the
Pushshift-API\footnote{\url{https://www.github.com/pushshift/api}}. We
only consider posts which have both text and an image.  From the
initial set of instances that we collected~(5,363) we manually removed
those with images of low quality, pornographic and sexually
inappropriate content, spam, or in a language other than English.

\subsection{Data Annotation}
We developed the annotation task with a subsample of 400 posts in a
preliminary experiment.  It was performed by two groups of three
students and with a direct interaction with the authors of this paper,
to obtain an understandable and unambiguous formulation of the
questions that we used for the actual crowdsourcing annotation.  The
actual annotation of 1,380~randomly sampled posts was then performed
with Amazon Mechanical Turk
(AMT\footnote{\url{https://www.mturk.com/}}) in two phases.  In the
first phase, we identify posts which likely contain an emotion by
asking
\begin{itemize}[topsep=0pt,itemsep=-1ex,partopsep=1ex,parsep=1ex]
\item [1.] Does the author want to express an emotion with the post?
\end{itemize}
In the second phase, we collect annotations for posts which
contain an emotion (we accept a post if 1/3 of the annotators marked
it as emotional) and ask
\begin{itemize}[topsep=0pt,itemsep=-1ex,partopsep=1ex,parsep=1ex]
\item [2.] What emotion did the author likely feel when writing this post?
\item [3.] What is the relation between the image and the text
    regarding emotion communication?
\item [4.] What is it in the image that triggers the emotion?
\end{itemize}
For both phases/experiments, we gather annotations by three
annotators. All questions allow one single answer. We show the
annotation interface on Amazon Mechanical Turk for the second phase
in the Appendix in Figure~\ref{fig:amtphase2}.

For the modelling which we describe in Section~\ref{sec:methods}, we
use a union of all labels from all annotators, acknowledging the
subjective nature of the annotation task. This leads to multi-label
classification, despite the annotation being a single-label annotation
task.

\noindent\textbf{Quality Assurance and Annotator Prescreening.}
Each potential annotator must reside in a predominantly
English-speaking country (Australia, Canada, Ireland, New Zealand,
United Kingdom, United States), and have an AMT approval rate of at
least~90\,\%.  Further, before admitting annotators to each annotation
phase, we showed them five manually selected posts that we considered
to be straightforward to annotate. For each phase, annotators needed
to correctly answer 80\,\% of the questions associated with those
posts.  Phase 1 had a 100\,\% acceptance rate; in Phase 2 this
qualification test had a 55\,\% acceptance rate. We summarize
participation and qualification statistics in
Tables~\ref{table:qualifications} and~\ref{table:patricipation} in the
Appendix.

\noindent\textbf{Annotators and Payment.}
Altogether, 75 distinct annotators participated in Phase 1, and 38
annotators worked in Phase 2. We paid \$0.02 for each post in Phase~1,
and \$0.08 for each post in Phase~2. The average time to annotate one
post was~16 and 38~seconds in Phase~1 and~2, respectively. This leads
to an average overall hourly wage of \$7. Overall, we paid \$337.44 to
annotators and \$105.06 for platform fees and taxes.

\subsection{Statistics of Annotated Dataset}
In total, 1,380 posts were annotated via AMT (we do not discuss the
preliminary annotations here). All results are summarized in
Table~\ref{tab:statistics}.
\begin{table}[t]
  \centering\small
  \setlength\tabcolsep{2.5mm}
  \begin{tabular}{llrrrr}
    \toprule
    & Label & $\ge1$ &  $\ge2$ & $=3$ & $\kappa$ \\
    \cmidrule(r){1-2}\cmidrule(rl){3-3}\cmidrule(rl){4-4}\cmidrule(lr){5-5}\cmidrule(l){6-6}
    \multirow{2}{*}{\rt{Emo.}}
    & Yes         & 1,061  & 670 & 333 & 0.3 \\
    & No          & 1,047 & 710 & 319 & 0.3 \\
    \cmidrule(r){1-2}\cmidrule(rl){3-3}\cmidrule(rl){4-4}\cmidrule(lr){5-5}\cmidrule(l){6-6}
    \multirow{8}{*}{\rt{Which emotion?}}
    &Anger        & 138 & 41  & 8   & .26 \\
    &Anticipation & 85  & 12  & 1   & .11 \\
    &Disgust      & 268 & 127 & 57  & .45 \\
    &Fear         & 64  & 15  & 5   & .28 \\
    &Joy          & 585 & 444 & 329 & .67 \\
    &Sadness      & 103 & 52  & 27  & .56 \\
    &Surprise     & 435 & 221 & 84  & .38 \\
    &Trust        & 54  & 6   & 1   & .11 \\
    &\it Overall      & \it 1732& \it 918 & \it 512 & \it .47 \\
\cmidrule(r){1-2}\cmidrule(rl){3-3}\cmidrule(rl){4-4}\cmidrule(lr){5-5}\cmidrule(l){6-6}
    \multirow{5}{*}{\rt{Relation?}}
    &Complementary & 1042 & 773 & 388 &    .02 \\
    &Decorative    & 124  & 6   & 0   &    .01 \\
    &Illustrative  & 476  & 152 & 4   &    .07 \\
    &Image only    & 142  & 27  & 0   &    .11 \\
    &Opposite      & 28   & 0   & 0   & $-$.01 \\
    &\it Overall       & \it 1812 & \it 958 &\it 392 & \it.04    \\
    \cmidrule(r){1-2}\cmidrule(rl){3-3}\cmidrule(rl){4-4}\cmidrule(lr){5-5}\cmidrule(l){6-6}
    \multirow{10}{*}{\rt{Stimulus?}}
    &Advertisement   & 23  & 4   & 0   & .14 \\
    &Animal          & 146 & 112 & 83  & .79 \\
    &Art/drawing     & 157 & 58  & 33  & .46 \\
    &Event/situation & 132 & 27  & 2   & .15 \\
    &Food            & 78  & 56  & 36  & .74 \\
    &Meme            & 129 & 58  & 8   & .34 \\
    &Object          & 211 & 102 & 51  & .50 \\
    &Person          & 260 & 168 & 91  & .61 \\
    &Place           & 46  & 12  & 5   & .34 \\
    &Screenshot      & 528 & 351 & 195 & .53 \\
    &\it Overall         & \it 1710& \it 948 &\it 504 &\it .53 \\
    \bottomrule
  \end{tabular}
  \caption{Corpus statistics for emotions, relations, and
    stimuli. ``$\ge1$", ``$\ge2$", ``$=3$" means that at least one, at
    least two, and all three annotators labeled the post with the
    respective emotion respectively. The overall number of posts
    that were annotated in Phase 1 is 1,380, and 1,054 for Phase
    2. $\kappa$ refers to Fleiss'
    kappa.}
  \label{tab:statistics}
\end{table}

\noindent\textbf{Did the author want to express an emotion
  with the post?} The total agreement of all three
annotators (=3) was achieved in~47\,\% of the time~(652 posts out of
1380). The
overall inter-annotator agreement for this question is fair, with
Fleiss~$\kappa$=.3. We consider this value to be acceptable for a
prefiltering step to remove clearly non-emotional posts for the actual
annotation in the next phase.

Of the 1,380 posts in Phase 1,~1,061 were labeled as ``emotion'', of
which seven were flagged as being problematic by annotators (see
Figure \ref{fig:amtphase2} in Appendix).  Therefore, in total, 1,054
posts are considered for Phase 2.

\noindent\textbf{What emotion did the author likely feel
  when writing this post?}
Table~\ref{tab:statistics} gives the individual counts of instances that received a particular
emotion label by at least one, two, or all three annotators. Note that the overall number of instances
can be greater than the number of instances in the case that
annotators disagree.  \textsl{Joy}, \textsl{surprise} and
\textsl{disgust} are the more frequent classes, with 585, 435, and 268
posts that received this label by at least one annotator. The number
of posts in which at least two annotators agreed is considerably
higher for \textsl{joy} than for the other emotions, which is also
reflected in the moderate overall inter-annotator agreement with
Fleiss $\kappa$=.47. For most classes, the agreement is moderate, with
some exceptions (\textsl{anger} is often conflated
with \textsl{disgust} as we will see below, and \textsl{anticipation},
and \textsl{trust}).

The agreement, however, can be considered to be similar to what has
been achieved in other (crowdsourcing-based) annotation studies. As
examples, \citet{purver-battersby-2012-experimenting} report an
agreement accuracy of 47\,\%. \citet{Schuff2017} report an agreement
of less than 10\,\% when a set of 6 annotators needed to label an
instance with the same emotion (but higher agreements for subsets of
annotators).

\noindent\textbf{What is the relation between the image and the
  text regarding emotion communication?}
The most dominant relations in our dataset are \textsl{complementary}
(1,042 instances in which one annotator decided for this label) and
\textsl{illustrative} (476).  There are fewer instances in which
annotators marked the relation \textsl{opposite} (28),
\textsl{decorative} (124) and that the text is not required to infer
the emotion (142).

The inter-annotator agreement is low, due to the skewness of the
dataset and a therefore high expected agreement: overall, we only
achieve $\kappa$=.04.  Note that this inbalanced corpus poses a
challenge in the results described in Section~\ref{sec:results}.

\noindent\textbf{What is it in the image that triggers the
  emotion?}
The emotion stimuli categories are more balanced: Most frequently,
people comment on what we classify as screenshots (528 out of 1054
received this label by at least one annotator), followed by depictions
of people (260), objects (211), pieces of art (157), and depictions
of animals (146). The agreement is moderate with an overall
$\kappa$=.53.  The labels \textsl{place} and \textsl{advertisement}
are underrepresented in the dataset.

\begin{figure}[t]
  \centering
  \includegraphics[width=1.0\linewidth]{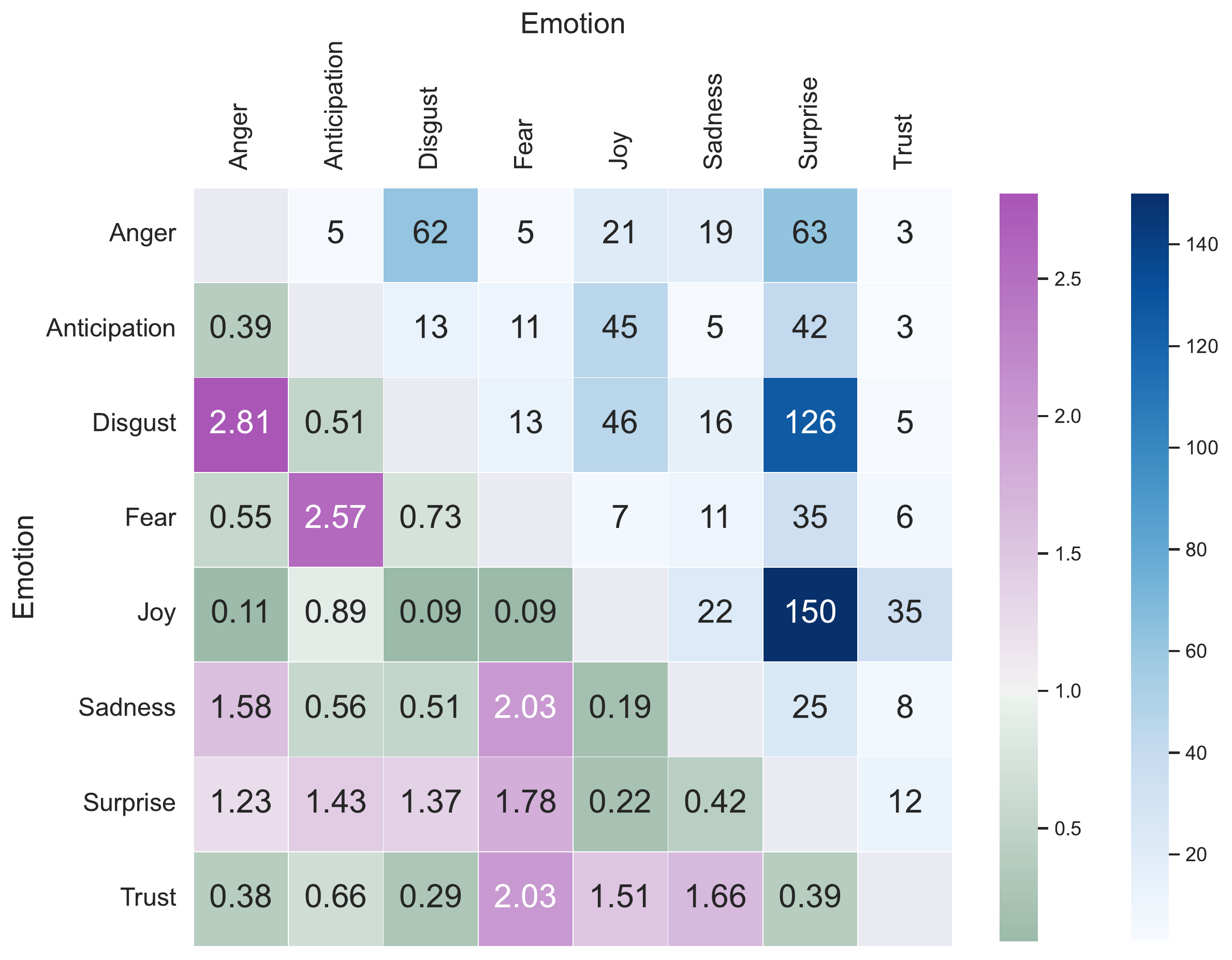}
  \caption{Emotion-Emotion cooccurrences. The values above the
    diagonal are absolute counts, while the numbers below the diagonal
    are odds ratios. I higher value denotes that the combination is
    particular specific.}
  \label{fig:emo-emo}
\end{figure}

\bigskip\noindent\textbf{Cooccurrences.}  We now turn to the question
which of the variables of the emotion category, the relation, and the
stimulus category cooccur.  Figure~\ref{fig:emo-emo} shows the results
with absolute counts above the diagonal, and odds-ratio values for the
cooccurrence of multiple emotions annotated by different annotators
below the diagonal \citep[details regarding the calculation can be
found in][]{Schuff2017}.  The emotion combinations of
\textsl{joy}--\textsl{surprise}~(150 times),
\textsl{surprise}--\textsl{disgust}~(126),
\textsl{surprise}--\textsl{anger}~(63), and
\textsl{disgust}--\textsl{anger}~(62 times) are most often used. This
is presumably an effect of the fact that people share information on
social media that they find newsworthy. Further, this shows the role
of \textsl{surprise} in combination with both positive and negative
emotions---as common in emotion annotations to limit ambiguity, we
modelled the task in a single-label annotation setup. Therefore, this
shows that different interpretations of the same post are possible.

The odds-ratio values point out the specificity of the combination of
\textsl{disgust}--\textsl{anger}. This could be explained with the
difference of these emotions regarding their motivational component,
namely to tackle a particular stimulus or to avoid it (known as the
fight-or-flight response). The combination of
\textsl{sadness}--\textsl{fear} can be explained with the importance
of the confirmation status of a stimulus (future or past) which
distinguishes these two emotions. This property might be ambiguous in
depictions in social media.  The
combinations of \textsl{fear}--\textsl{anticipation} and
\textsl{fear--trust} might be considered surprising. Such combinations
of positive and negative emotions frequently occur in motivational
text depictions, for instance ``don't be afraid of
your fears''.

\begin{figure*}[t]
  \centering
  \subfloat[][Counts]{\includegraphics[width=7.5cm]{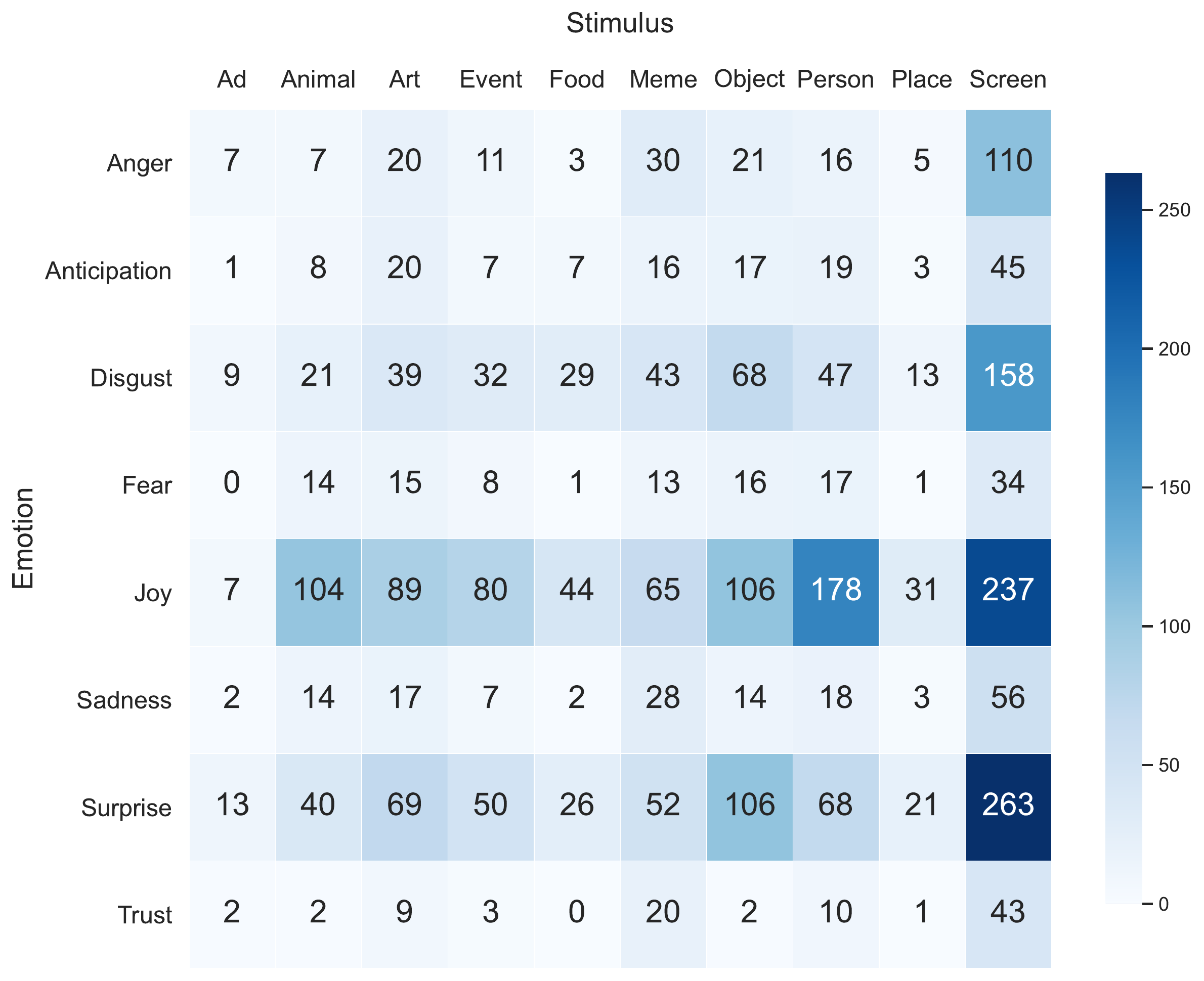}}\hspace{5mm}
  \subfloat[][Odds Ratio]{\includegraphics[width=7.5cm]{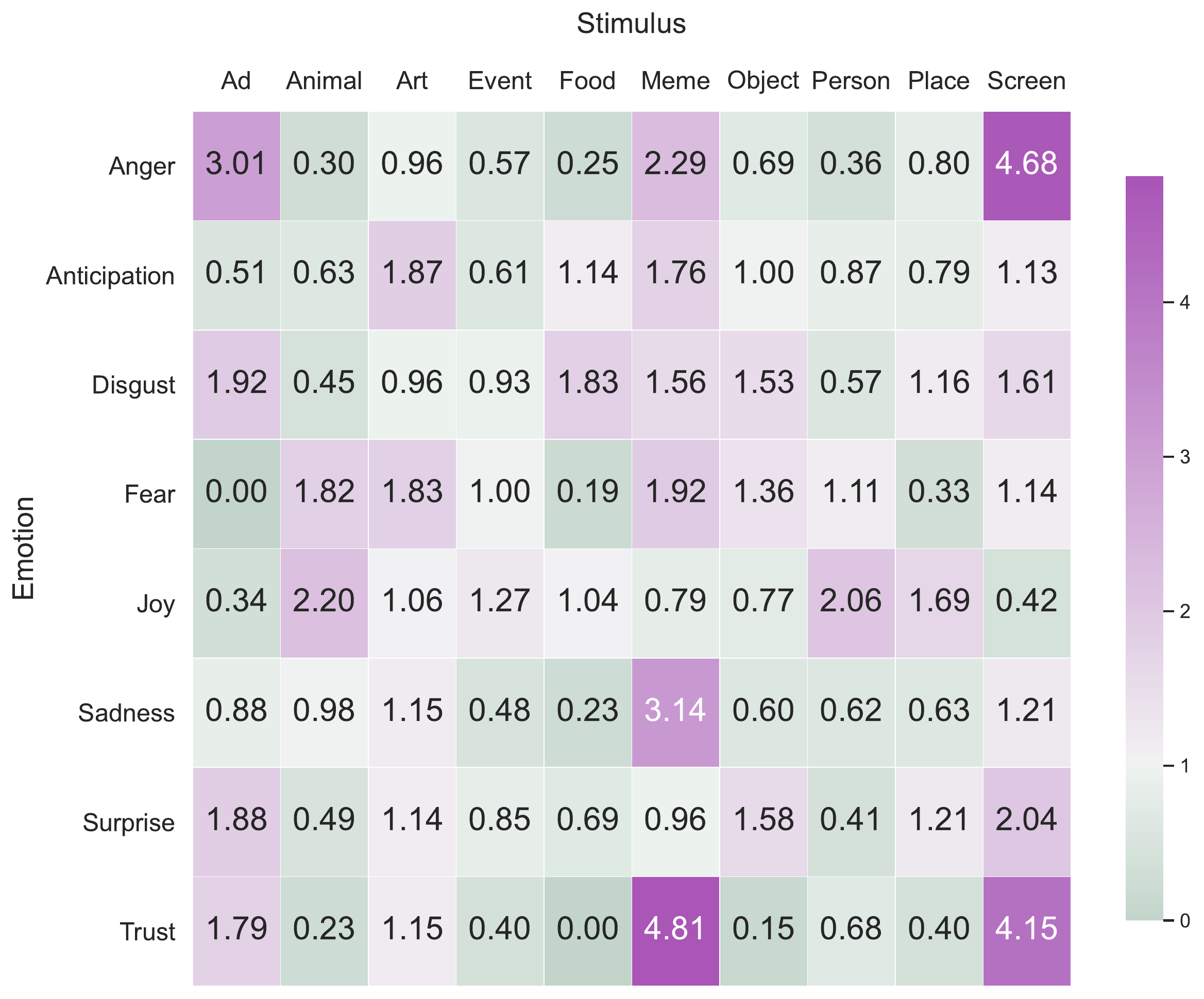}}
  \caption{Emotion-Stimulus Cooccurrences.}
  \label{fig:stim-emo}
\end{figure*}

We show the cooccurrence counts and odds ratios for the stimulus and
the emotion in Figure~\ref{fig:stim-emo}. For the emotions
\textsl{anger}, the stimuli of \textsl{advertisments} and
\textsl{screenshots} are outstanding. \textsl{Anticipation} has the
highest value for art. \textsl{Disgust} is particularly specific for
\textsl{food} and \textsl{advertisement}. This shows the metaphoric
use of the term (in the sense of repugnance) and a more concrete use
(in the sense of revulsion). Interestingly, \textsl{fear} is specific
for stimuli of \textsl{animals}, \textsl{art}, and
\textsl{memes}. \textsl{Joy} is the only emotion that has a high odds
ratio with \textsl{places}, and \textsl{persons}, but also with
\textsl{animals}. \textsl{Sadness} and \textsl{trust} have the highest
value for \textsl{memes}.

We do not discuss the \textsl{relation} category further, given the predominance 
of the \textsl{complementary} class and its limited
inter-annotator agreement.

\section{Methods}
\label{sec:methods}
In the following, we present the models that we used to predict (1)
each variable (emotion, stimulus, relation) separately in each
modality, and (2) across modalities with joint models.

\subsection{Text}
For the text-based model, we fine-tune the pre-trained RoBERTa
model\footnote{\hbox{\url{https://huggingface.co/transformers/model_doc/roberta.html}}}
\citep{Liu:2019roberta}. We perform multi-task learning for emotion,
stimulus and relation by adding a fully connected layer (for each set of labels), on top of
the last hidden layer. The model combines the loss for all three sets
of labels and updates the weights accordingly during the training
phase.\footnote{Our first choice of only one layer performed en par to
  multiple stacked layers.} We use a learning rate of
\mbox{$3\cdot 10^{-5}$} for all layers, except for the top three fully
connected ones \mbox{($3\cdot 10^{-3}$)}. We use the learning rate
scheduler with a step size of 5 and train for maximally 20 epochs, but
perform early stopping if the validation loss does not improve by more
than 0.005\%.

\subsection{Images}
We built the image-based model on top of a pretrained deep residual
network model with~48 convolutional layers and~2 pooling layers
\cite[ResNet,~][]{He:2016resnet}.  We use the ResNet50 that is
provided by
PyTorch\footnote{\url{https://pytorch.org/hub/pytorch_vision_resnet/}}
and was pretrained on 1,000~ImageNet categories
\cite{ILSVRC15,imagenet_cvpr09}.  As with the text-based model, we add
three fully connected layers on top of the fully connected layer of
the ResNet50 model, with the sigmoid activation function. Unlike
RoBERTa, we do not fine-tune the convolutional layers to prevent the
pre-trained weights to change.\footnote{We performed experiments with
  unfreezing several top convolutional layers, however, it did not
  lead to better results.}

\subsection{Joint Models}
We evaluate three simple multimodal methods which combine the information
from the text and the image modality on the traditional three different stages:  
early, late, and model-based fusion \citep{Snoek:2005fusions}. 

In
early (feature-based) fusion, the features extracted from both
modalities are fused at an early stage and passed through a
classifier. As the input, our early-fusion model takes the tokenized
text and preprocessed image (images are resized, converted to tensors,
and normalized by the mean and standard
deviation\footnote{\url{https://pytorch.org/vision/stable/transforms.html}}),
and concatenates them into one vector to pass through the final
classifier, that consists of several layers (three linear, dropout,
and three fully connected layers) with the input size depending on the
longest text in the training set and output size depending on the
task. The activation function is, as in all our models, a sigmoid
function.
    
In late (decision-based) fusion, classification scores are obtained
for each modality separately.  These scores are then fed into the
joint model. In our late-fusion model, we pass the text and image
through the text-based and image-based models respectively, and
concatenate the output probabilities of these
models.\footnote{Experiments with summed vectors did not improve
  results.}  We then pass this vector through a fully connected layer
with twice the number of classes from the two models as input and
output, and apply sigmoid for prediction.  That is, for the emotion
classification, the vectors of eight labels from RoBERTa and ResNet50,
summing up to 16, are passed to the fully connected layer.
    
For model-based fusion, we extract text and image features from our
unimodal text and image-based classifiers, respectively (from the last
hidden layers before the fully conntected ones), and feed these to a
final classifier.\footnote{Experiments with more complex models with
  multiple top layers did not improve results, thus, we chose a
  single-layer-on-top model for the experiments.}

\section{Results}
\label{sec:results}
We evaluate our models on predicting emotions, text--image relations,
and emotion stimuli using unimodal and multimodal models, based on the
\F measure. We use the dataset of 1054 instances in which we
aggregate the labels from the three annotators by accepting a label if
one annotator assigned it (this approach might be considered a
``high-recall'' aggregation of the labels, similar to
\citet{Schuff2017}). Despite being a single-label annotation task,
this leads to a multi-label classification setup. In other words, the annotation process requires annotators to select a single label (for each set of labels), e.g. one emotion per post; however, the experiments are conducted using multiple labels per set, depending on how many labels are given by three annotators for each set of labels. The data is randomly
split into 853 instances for training, 95 instances for validation,
and 106 test instances.

\begin{table}[t]
  \centering\small
  \renewcommand{\arraystretch}{1.2}
  \begin{tabular}{llccc}
    \toprule
    & Model & Emo. & Rel. & Stim. \\
    \cmidrule(rl){2-2}\cmidrule(rl){3-3}\cmidrule(lr){4-4}\cmidrule(l){5-5}
          & Majority Baseline  & .22          & .56          & .21 \\
    \cmidrule(rl){2-2}\cmidrule(rl){3-3}\cmidrule(lr){4-4}\cmidrule(l){5-5}
    \multirow{2}{*}{\parbox{10mm}{uni-modal}}
          & Text               & \textbf{.53} & \textbf{.77} & .45 \\
          & Image              & .41          & .67          & .59 \\
    \cmidrule(rl){1-2}\cmidrule(rl){3-3}\cmidrule(lr){4-4}\cmidrule(l){5-5}
    \multirow{3}{*}{\parbox{10mm}{multi-modal}}
          & Early fusion       & .40          & .72          & .33 \\
          & Late fusion        & .47          & .72          & .41 \\
          & Model-based fusion & \textbf{.53} & .76          & \textbf{.63} \\
    \bottomrule
  \end{tabular}
  \caption{Experimental results in predicting emotions, relations, and
    stimuli using unimodal and multimodal models. The results are
    presented in weighted \F score. Bold face indicates the highest
    value in each column/task.
  }
  \label{tab:resultssummary}
\end{table}

Table~\ref{tab:resultssummary} summarizes the results, averaging
across the values for each class variable. We observe that the
emotions and the relations can be predicted with the highest \F with
the text-based unimodal model. The discrepancy to the image-based
model is substantial, with~.53 to~.41 for the emotions and~.77 to~.67
for the relations. The stimulus detection benefits from the multimodal
information from both the image and the text---the highest
performance, .63, is achieved with the model-based fusion
approach. From the unimodal models, the image-based model is performing
better than the text-based model. This is not surprising---in
multimodal social media posts that express an emotion, the depictions
predominantly correspond to a stimulus, or their identification is at
least important. The corpus statistics show that: posts in which the
image is purely used decoratively are the minority.

\begin{table}[t]
  \centering\small
  \renewcommand{\arraystretch}{1.1}
  \begin{tabular}[t]{llrrrrr}
    \toprule
    && \multicolumn{2}{c}{Unimodal} & \multicolumn{3}{c}{Multimodal} \\
    \cmidrule(r){3-4}\cmidrule(l){5-7}
    & Label & Txt & Img & Early & Late & Mb.\\
    \cmidrule(r){2-2}\cmidrule(lr){3-3}\cmidrule(rl){4-4}\cmidrule(rl){5-5}\cmidrule(rl){6-6}\cmidrule(l){7-7}
    \multirow{8}{*}{\rt{Emotions}}
    &Anger           & .08  & .04 & .03  & 0    & \textbf{.14} \\
    &Anticipation    & 0    & 0   & \textbf{.12}  & 0    & 0 \\
    &Disgust         & \textbf{.47}  & .23 & .26  & .27  & .39 \\
    &Fear            & 0    & 0   & \textbf{.04}  & 0    & 0 \\
    &Joy             & \textbf{.84}  & .66 & .64  & .85  & .78 \\
    &Sadness         & .28  & 0   & .09  & 0    & \textbf{.37} \\
    &Surprise        & .61  & .57 & .52  & .57  & \textbf{.70} \\
    &Trust           & \textbf{.04}  & 0   & 0    & 0    & 0 \\
    \hline
    \multirow{5}{*}{\rt{Relations}}
    &Compl.          & \textbf{.99}  & \textbf{.99} & .98  & \textbf{.99} & \textbf{.99} \\
    &Decorative      & .05  & 0   & .11  & 0    & \textbf{.20} \\
    &Illustrative    & .65  & .46 & .50  & \textbf{.66} & .61 \\
    &Image-only      & \textbf{.38}  & 0   & .24  & 0    & .34 \\
    &Opposite        & 0    & 0   & 0    & 0    & 0 \\
    \hline
    \multirow{10}{*}{\rt{Stimuli}}
    &Advert.         & 0    & 0   & 0    & 0    & 0    \\
    &Animal          & .60  & \textbf{.78} & .28  & .52 & .74 \\
    &Art/drawing     & .02  & .27 & .26  & 0    & \textbf{.45} \\
    &Event/sit.      & .20  & .16 & .13  & 0    & \textbf{.29} \\
    &Food            & .22  & \textbf{.63} & .19  & 0    & .62 \\
    &Meme            & .40  & .35 & .17  & 0    & \textbf{.57} \\
    &Object          & .32  & \textbf{.68} & .17  & .25 & .65 \\
    &Person          & .43  & \textbf{.59} & .21  & .52 & \textbf{.59} \\
    &Place           & 0    & .06 & .04  & 0    & .19 \\
    &Screenshot      & .72  & .78 & .63  & .78 & \textbf{.79} \\
    \bottomrule
  \end{tabular}
  \caption{Experimental results for all labels in predicting emotions,
    relations, and stimuli using the text-based and image-based
    unimodal models, and fusion models. The results are presented in F1 score.}
  \label{tab:detailedresults}
\end{table}

Table~\ref{tab:detailedresults} shows detailed per-label results.  For
the \textbf{emotion classification} task, we see that for three emotions, the
text-only model leads to the best performance (\textsl{disgust},
\textsl{joy}, \textsl{trust}, while the latter is too low to draw a
conclusion regarding the importance of the modalities). The other
emotions benefit from a multimodal approach. Overall, still, the
text-based model shows highest average performance, given the
dominance of the emotion \textsl{joy}.

For most \textbf{stimulus} categories, either the image or a
multimodal model performs best. This is not surprising, given that the
stimulus is often depicted in the visual part of a multimodal
post. More complex depictions that could receive various evaluations,
like \textsl{art}, \textsl{events/situations}, and \textsl{memes}
require multimodal information. In those, the image information alone
is not sufficient---the performance difference is between 22pp and
13pp in \F. For those stimuli, in which the text-based model
outperforms the multimodal models, the difference is lower. The
text-based model is never performing best, but shows acceptable
performance for \textsl{animals}, \textsl{memes}, \textsl{screenshots}
and \textsl{person} depictions.

Regarding the \textbf{relations}, the \textsl{complementary} class is
predicted with the best performance; which is due to the frequency of
this class. The label \textsl{decorative} can only be predicted with a
(slightly) acceptable performance with the multimodal approach, while
\textsl{illustrative} predictions based on text-only are nearly en par
with a multimodal model.

From the three multimodal fusion approaches, early fusion performs the
worst, followed by late fusion. Model-based fusion most often leads to
the best result.  We show examples for instances in which the
multimodal model performs better than unimodal models in Table
\ref{tab:discussion} in the Appendix.

\section{Conclusion and Future Work}
With this paper, we presented the first study on how users in social
media make use of text and images to communicate their emotions. We
have seen that the number of multimodal posts in which the image does
not contribute additional information over the text is in the
minority, and, hence, interpretation of images in addition to the text
is important.  While the inter-annotator agreement for relation was
not reliable enough to draw this conclusion, prediction of stimulus
correlates with prediction of emotion due to the information that is
present in the image but missing in the text, and thus makes images
play a significant role in analysis of social media posts. This is
also the first study on stimulus detection in multimodal posts, and we
have seen that for the majority of stimulus categories, the
information in the text is not sufficient.

In contrast to most work on emotion stimulus and cause detection in
NLP, we treated this task as a discrete classification task, similar
to early work in targeted sentiment analysis. An interesting step in
the future will be to join segment-based open domain stimulus
detection, as it is common in text analysis, with region-based image
analysis, and ground the textual references in the image. This will
allow to go beyond predefined categories.

\section*{Acknowledgements}
This work was supported by Deutsche Forschungsgemeinschaft (project
CEAT, KL 2869/1-2).

\bibliography{lit}
\bibliographystyle{acl_natbib}

\clearpage

\appendix

\onecolumn

\section{Appendix}
\label{sec:appendix}

\begin{figure*}[h!]
      \centering\sffamily\footnotesize

      \subfloat[][%
      Person/people
      ]{%
        \fbox{%
          \begin{minipage}[t][4cm]{0.17\linewidth}
            I lost my smile for a while. Just felt happy today first time in a long time.\\
            \includegraphics[height=2.8cm]{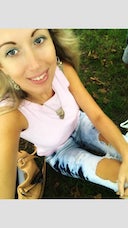}
          \end{minipage}
        }
      }
      \hfill
      \subfloat[][%
      Animal
      ]{%
        \fbox{%
          \begin{minipage}[t][4cm]{0.17\linewidth}
            have wanted them for 40 years - they arrived today. meet harvey and cooper.\\
            \includegraphics[height=1.8cm]{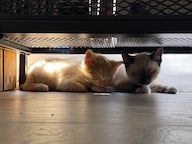}
          \end{minipage}
        }
      }
      \hfill
      \subfloat[][%
      Object
      ]{%
        \fbox{%
          \begin{minipage}[t][4cm]{0.17\linewidth}
            So, I am turning 23 and found out I am good at chess. Never to late to pick up a new hobby.\\
            \includegraphics[height=2cm]{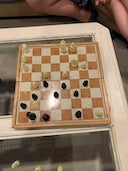}
          \end{minipage}
        }
      }
      \hfill
      \subfloat[][%
      Food
      ]{%
        \fbox{%
          \begin{minipage}[t][4cm]{0.17\linewidth}
            This spaghetti after it sat in a bowl for a night\\
            \includegraphics[height=1.8cm]{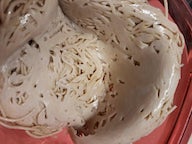}
          \end{minipage}
        }
      }
      \hfill
      \subfloat[][%
      Meme
      ]{%
        \fbox{%
          \begin{minipage}[t][4cm]{0.17\linewidth}
            Noooooooo\\
            \includegraphics[height=3cm]{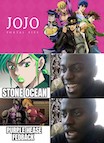}
          \end{minipage}
        }
      }

      \subfloat[][%
      Screenshot/text in image
      ]{%
        \fbox{%
          \begin{minipage}[t][4cm]{0.17\linewidth}
            How expensive is coffee where they live??\\
            \includegraphics[height=2.8cm]{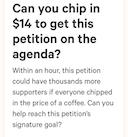}
          \end{minipage}
        }
      }
      \hfill
      \subfloat[][%
      Art/drawing
      ]{%
        \fbox{%
          \begin{minipage}[t][4cm]{0.17\linewidth}
            Why did I make this?\\
            \includegraphics[height=1.8cm]{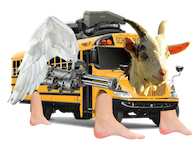}
          \end{minipage}
        }
      }
      \hfill
      \subfloat[][%
      Advertisement
      ]{%
        \fbox{%
          \begin{minipage}[t][4cm]{0.17\linewidth}
            what the fuck is this vpn for \\
            \includegraphics[height=1.8cm]{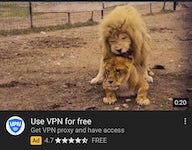}
          \end{minipage}
        }
      }
      \hfill
      \subfloat[][%
      Event/situation
      ]{%
        \fbox{%
          \begin{minipage}[t][4cm]{0.17\linewidth}
            Women had their first ever pro wrestling match in Saudi Arabia\\
            \includegraphics[height=1.4cm]{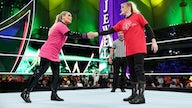}
          \end{minipage}
        }
      }
      \hfill
      \subfloat[][%
      Place
      ]{%
        \fbox{%
          \begin{minipage}[t][4cm]{0.17\linewidth}
            portland japanese garden\\
            \includegraphics[height=1.7cm]{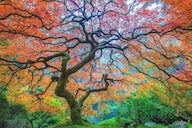}        
          \end{minipage}
        }
      }
      \caption{Examples of emotion stimuli in post images.}
      \label{fig:example3}
    \end{figure*}

\begin{figure*}[h!]
  \centering
  \fbox{\includegraphics[width=\textwidth]{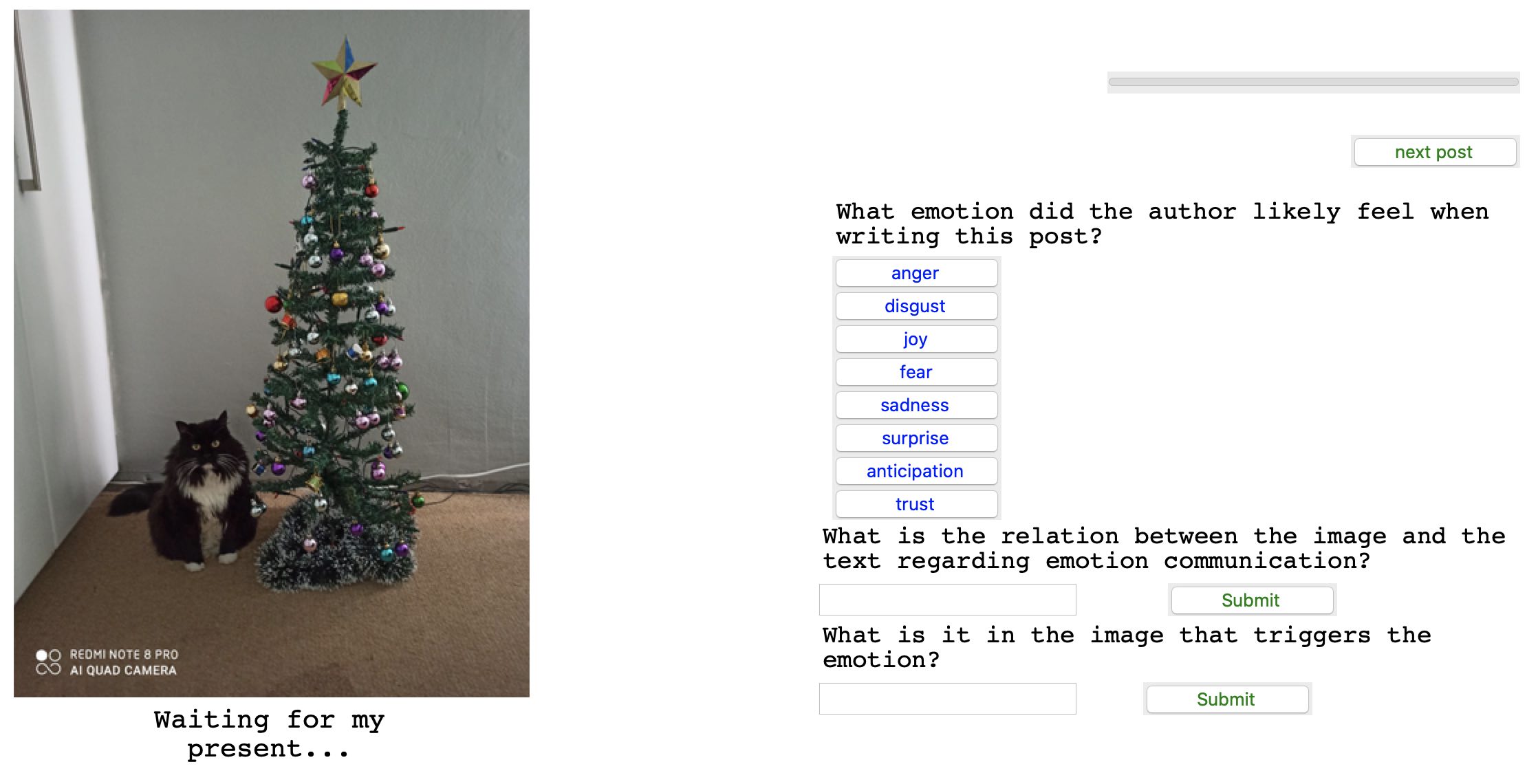}}
  \caption{Annotation tool to define taxonomies.} \label{fig:tool}
\end{figure*}

\begin{figure}[h!]
  \centering
    \fbox{\includegraphics[width=\textwidth]{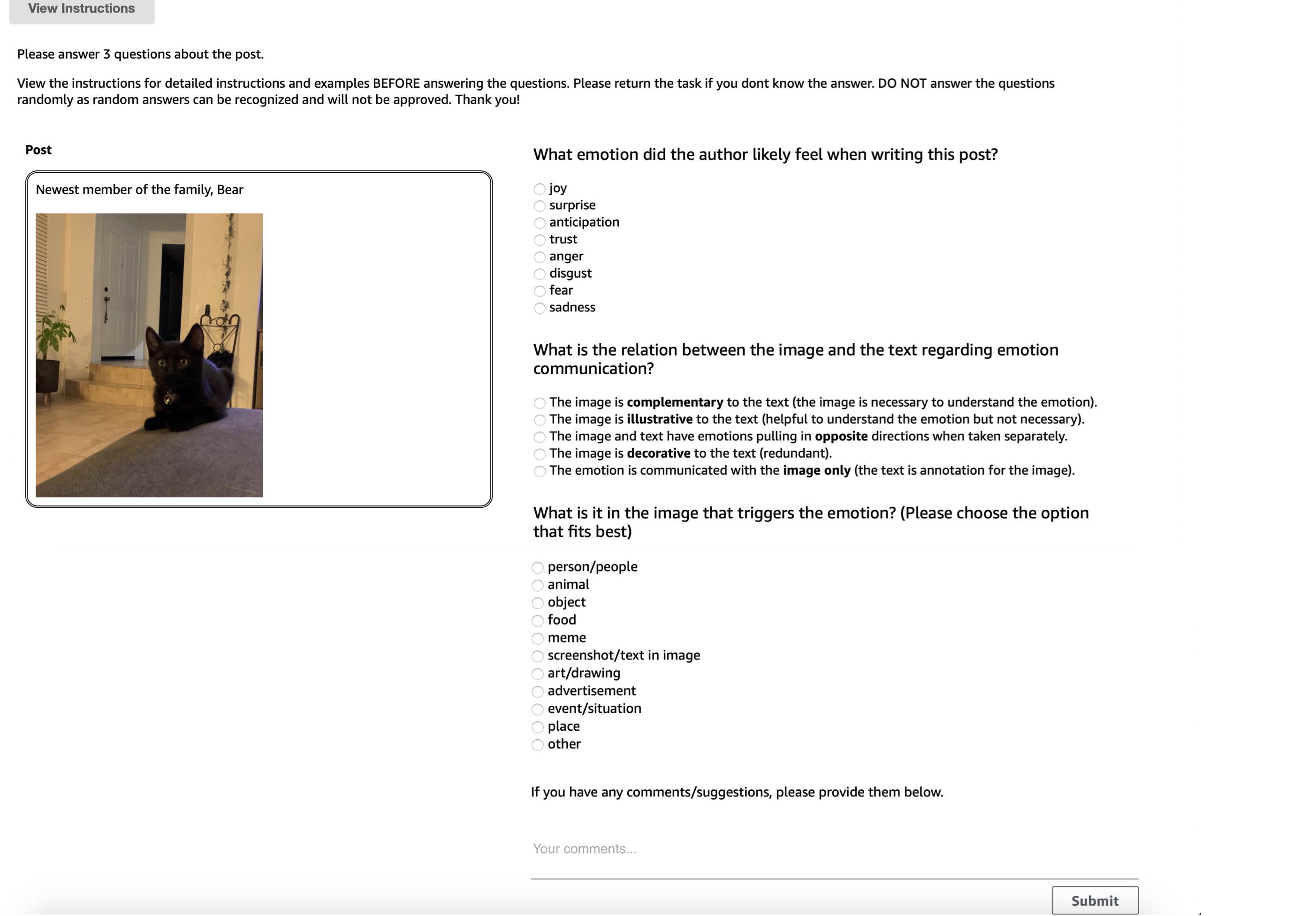}}
  \caption{Annotation Environment on Amazon Mechanical Turk}
  \label{fig:amtphase2}
\end{figure}

\begin{table*}[h!]
  \centering
  \begin{tabular}{| p{0.35\linewidth} | p{0.6\linewidth} |}
    \hline
    \textbf{Qualification} & \textbf{Description} \\
    \hline
    {Qualification test 1: emotional/non-emotional posts} & {5 posts presented to annotators to label the post emotional or non-emotional; passing score of 80\%} \\
    \hline
    {Qualification test 2: emotion, relation, stimulus identification} & {5 posts presented to annotators to label the post for emotions, relations, and stimuli; passing score of 80\%} \\
    \hline
    {Region} & {Annotators must reside in either of the six English-speaking countries (Australia, Canada, Ireland, New Zealand, United Kingdom, United States) to force the task to be done by native speakers.}\\
    \hline
    {Human Intelligence Task (HIT) approval rate} & {The HIT approval rate represents the proportion of completed tasks that are approved by Requesters and ensures the quality of the job workers do on the platform.}\\
    \hline
  \end{tabular}
  \caption{Qualifications used on AMT for data annotation.}
  \label{table:qualifications}
\end{table*}
    
\begin{table*}[h!]
  \centering
  \begin{tabular}{|l|r|r|r|r|}
    \hline
    {} & \multicolumn{2}{c|}{{Qualification}} & \multicolumn{2}{c|}{{Participation}} \\
    \cline{2-3} \cline{4-5}
    {} & Attempted & Passed & From previous task & New \\
    \hline
    Task 1 & 75 & 75  & - & 75 \\
    Task 2 & 69 & 38  & 17 & 21 \\
    \hline
  \end{tabular}
  \caption{Statistics on participation for the two tasks. All numbers
    are the counts of workers. Qualification tests are described in
    Table \ref{table:qualifications}. \textsl{Attempted} are the number of workers that took the qualification test, while \textsl{passed} is the number of workers that answered at least 80\% of the questions correctly. \textsl{From previous task} refers to the number of workers that participated in Phase 1 as well as Phase 2, while \textsl{new} are the participants that have not participated in the previous phase.}
  \label{table:patricipation}
\end{table*}

\begin{table*}[h!]
  \centering
\begin{tabularx}{\linewidth}{lXlcccc}
  \toprule
  &&&& \multicolumn{3}{c}{Predictions (Emotions/Stimulus)} \\
  \cmidrule(r){5-7}
  &Text & Image & Gold & Image-only & Text-Only & Multimodal \\
  \cmidrule(rl){1-2}\cmidrule(r){3-3}\cmidrule(rl){4-4}\cmidrule(lr){5-5}\cmidrule(lr){6-6}\cmidrule(l){7-7}
  \multirow{17}{*}{\rt{Emotion}}
  & Found a fly in my tea halfway through it 
  & \vertimg{0.3}{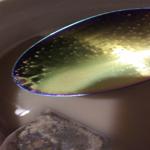} 
  & Disgust 
  & Joy   
  & Disgust/Surprise   
  & Disgust
  \\
  \cmidrule(rl){2-2}\cmidrule(r){3-3}\cmidrule(rl){4-4}\cmidrule(lr){5-5}\cmidrule(lr){6-6}\cmidrule(l){7-7}
  & Dont know if it has been posted before but here u go 
  & \vertimg{0.3}{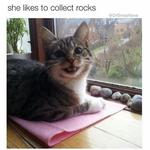} 
  & Joy 
  & Joy   
  & Disgust   
  & Joy
  \\
  \cmidrule(rl){2-2}\cmidrule(r){3-3}\cmidrule(rl){4-4}\cmidrule(lr){5-5}\cmidrule(lr){6-6}\cmidrule(l){7-7}
  & I find a monster under my bed
  & \vertimg{0.3}{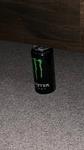} 
  & Sadness 
  & Joy   
  & Fear/Surprise   
  & Surprise
  \\
  \cmidrule(rl){2-2}\cmidrule(r){3-3}\cmidrule(rl){4-4}\cmidrule(lr){5-5}\cmidrule(lr){6-6}\cmidrule(l){7-7}
  & Definitely stoked with how much weight I’ve lost since overcoming my alcoholism!
  & \vertimg{0.3}{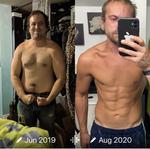} 
  & Joy 
  & Fear
  & Joy   
  & Joy
  \\
  \cmidrule(rl){1-2}\cmidrule(r){3-3}\cmidrule(rl){4-4}\cmidrule(lr){5-5}\cmidrule(lr){6-6}\cmidrule(l){7-7}
  \multirow{20}{*}{\rt{Stimulus}}
  & It causes unnatural amounts of pain to just look at it 
  & \vertimg{0.3}{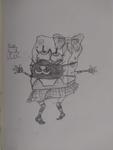} 
  & Art/Drawing  
  & Art/Drawing
  & Person
  & Art/Drawing
  \\
  \cmidrule(rl){2-2}\cmidrule(r){3-3}\cmidrule(rl){4-4}\cmidrule(lr){5-5}\cmidrule(lr){6-6}\cmidrule(l){7-7}
  & I have no idea the context of this picture from Steam Powered Giraffe but the sheer happiness in it makes me happy. Hope it does for you, too! You see, he never smiles that big! 
  & \vertimg{0.3}{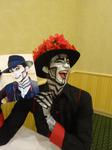} 
  & Person  
  & ---
  & Person
  & Person
  \\
  \cmidrule(rl){2-2}\cmidrule(r){3-3}\cmidrule(rl){4-4}\cmidrule(lr){5-5}\cmidrule(lr){6-6}\cmidrule(l){7-7}
  & After multiple tries, my sunflower finally bloomed! What a beauty.
  & \vertimg{0.3}{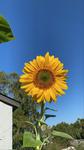}
  & Object
  & ---
  & ---
  & Object
  \\
  \bottomrule
\end{tabularx}
\caption{Examples in which the multimodal model-based model returns
  the correct result, but at least one unimodal model does
  not. ``---'' means that the model was not confident enough to predict any of the labels from the set.}
  \label{tab:discussion}
\end{table*}

\end{document}